\documentclass[preprint,12pt]{elsarticle}
\makeatletter

\def\ps@pprintTitle{%

  \let\@oddhead\@empty

  \let\@evenhead\@empty

  \let\@oddfoot\@empty

  \let\@evenfoot\@empty

}

\makeatother

\usepackage{amssymb}
\usepackage{amsmath}

\usepackage{url}
\usepackage{xcolor}
\usepackage{ulem}

\usepackage{subcaption}

\begin{document}

\begin{frontmatter}



\title{Parallel Training Using a CNN-DNN Architecture
for Accelerated Development of Diagnostic Models}


\author[label1,label2]{Janine Weber-Hamacher\fnref{label11}} 

\author[label3]{Astha Jaiswal\fnref{label11}} 

\author[label3]{Philipp Fervers} 

\author[label4,label16]{Dorottya Móré} 

\author[label4,label15]{Athanasios Giannakis} 

\author[label14]{Ricarda Fischbach} 

\author[label5]{Andreas Michael Bucher} 

\author[label3]{Rahil Shahzad}

\author[label3]{Jonathan Kottlors} 

\author[label3]{Thorsten Persigehl\fnref{label12}} 

\author[label1,label2]{Axel Klawonn\fnref{label12}} 

\fntext[label11]{Shared first authorship}

\fntext[label12]{Shared last authorship}

\affiliation[label1]{organization={Department of Mathematics and Computer Science, University of Cologne},
            city={Cologne}, 
            country={Germany}}

\affiliation[label2]{organization={Center for Data and Simulation Science, University of Cologne},
            country={Germany}}

\affiliation[label3]{organization={Institute for Diagnostic and Interventional Radiology, Faculty of Medicine and University Hospital Cologne, University of Cologne},
            city={Cologne}, 
            country={Germany}}
            
\affiliation[label4]{organization={Department of Diagnostic and Interventional Radiology, University Hospital Heidelberg, University of Heidelberg},
            city={Heidelberg}, 
            country={Germany}}

\affiliation[label5]{organization={Institute for Diagnostic and Interventional Radiology, Frankfurt University Hospital},
            city={Frankfurt}, 
            country={Germany}}
            
\affiliation[label14]{organization={Department of Paediatrics, University Hospital Mainz},
            city={Mainz}, 
            country={Germany}}
            
\affiliation[label15]{organization={2nd Department of Radiology, Attikon University General Hospital},
            city={Athens}, 
            country={Greece}}
            
\affiliation[label16]{organization={Department of Radiology and Nuclear Medicine, University Hospital Basel},
            city={Basel}, 
            country={Switzerland}}
            
\begin{abstract}
 Artificial intelligence has shown promise in assisting radiologists in imaging-based diagnosis across a wide range of diseases. Efficient training of large deep learning models is essential to cope with extremely large data sets or dynamically growing disease data, like in a pandemic like situation. In this retrospective study, we collected 300 CT scans from COVID-19 and non-COVID-19 pneumonia patients from three different centers in Germany. We investigated a hybrid CNN-DNN network model based on image decomposition and localization that naturally supports parallel and efficient training of deep learning models. In total, 156 models with three different architectures were trained to capture features at different levels resulting in 12 patient-level COVID-19 diagnosis models. Diagnostic performance as well as time saving were measured. 
 The highest accuracy was obtained from DenseNet121 and 3D CNN models with a parallel CNN-DNN approach, resulting in $88.78\%$ training, $76.67\%$ validation and $76.03\%$ test accuracy for the DenseNet121 with $4\times4\times1$ subdomains and $87.72\%$ training, $76.82\%$ validation and $74.86\%$ test accuracy, respectively, for the 3D CNN with $4\times4\times1$ subdomains. The strongest reduction in parallel training time by a factor of $31$ was observed for the 3D CNN model and $4\times4\times2$ subdomains.
 Our parallel training approach improves efficiency as well as performance enabling rapid model development, among others crucial for pandemic preparedness.
\end{abstract}


%

\end{frontmatter}


\section*{Abbreviations}

\begin{table}[h!]
\begin{tabular}{ll}
AI & Artificial intelligence \\
ML & Machine learning \\
CNN & Convolutional neural network \\
COVID-19 & Coronavirus disease 2019 \\
DNN & Dense feedforward neural network \\
nCP & Non-COVID-19 pneumonia \\
ResNet & Residual network \\
DenseNet & Densely connected neural network \\
ROC & Receiver operating characteristic curve \\
AUC & Area under curve
\end{tabular}
\end{table}

\section{Introduction}
\label{sec:intro}
Medical images enable non-invasive assessment of health conditions allowing diagnosis, monitoring, tracking and incidental finding of various diseases such as brain tumors~\cite{ghasemi2025detection}, ovarian tumors~\cite{jan2023machine}, hepatocellular carcinoma~\cite{zhang2024ct}, interstitial lung disease~\cite{zhang2025deep}, respiratory disease~\cite{geroski2024softlungx}, cirrhosis~\cite{zheng2024fully}, usual interstitial pneumonia~\cite{chung2024deep}, or opportunistic screening of osteoporosis~\cite{huang2025application}. Artificial intelligence (AI) algorithms streamline radiology workflows and enhance diagnostic sensitivity ~\cite{avakian2026artificial}. AI based disease diagnosis is not only important for improved routine patient care, and to deal with workforce shortage ~\cite{jing2025ai} but critical to effectively deal with emergency and pandemic like situations that massively burden the health care system and where a prompt response is crucial.

AI is a rapidly developing field in radiology with new models, deep neural network architectures, and rapidly growing public data sets for training these models. Thus, efficient training methods are needed to find the best performing AI model and to fine tune the most promising models for the specific indications. In this study, we use coronavirus disease 2019 (COVID-19) as an example use-case. The COVID-19 pandemic has shown that for new diseases with unknown dynamics and therapy, prompt response of health care service in diagnosis and isolation is essential ~\cite{inui2021role, fervers2022assessment, rajpoot2024integrated, farahat2024ai}. To deal with rapidly and dynamically growing disease data, efficient training of imaging-based deep diagnostic models ~\cite{pham2025leveraging, wang2021deep, baghdadi2022automated} is crucial for pandemic preparedness~\cite{jaiswal2025}.

In general, obtaining high accuracy values in AI-assisted medical image recognition often results in Machine learning (ML) models with millions or billions of parameters and hence, in long training times. Thus, efficient parallelization approaches for the fast training of such models have become of crucial importance; see also~\cite{ben2019demystifying} and~\cite{klawonn2024survey} for an overview of various parallelization methods of ML and, in particular, neural network models. 

\begin{figure}[ht]
    \centering
    \includegraphics[width=0.95\textwidth]{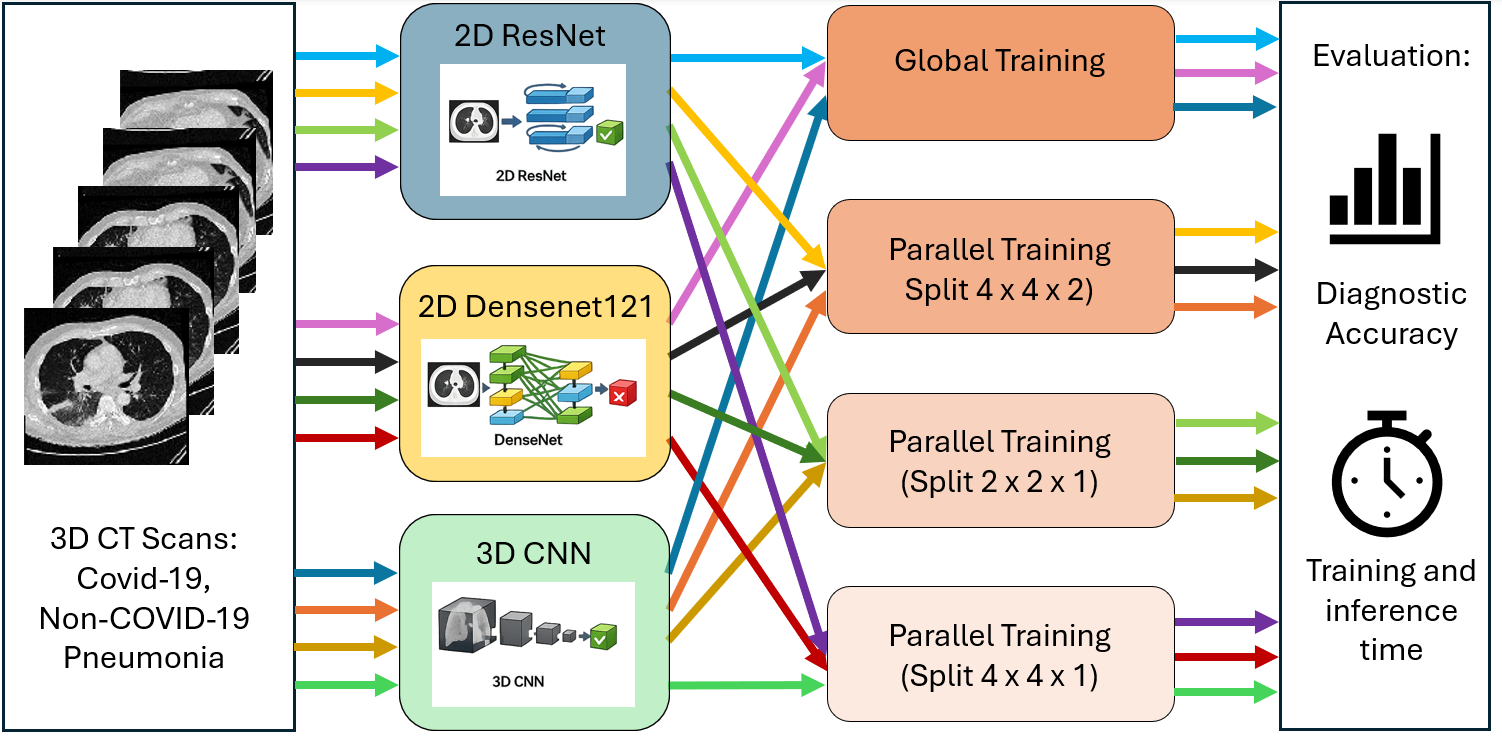}
    \caption{Experimental design of this study.  Each experiment workflow is shown in a different arrow color.}
    \label{fig:ex_design}
\end{figure}

In this work, we focus on a concrete model parallel training strategy for the parallel training of ML models for the classification of image data. The considered method was initially proposed in~\cite{klawonn2024domain} where it was applied to different synthetic and open source datasets. Especially for the classification of three-dimensional image data, that is, the chest CT scans, by a convolutional neural network (CNN) with three-dimensional filters, the model parallel method in~\cite{klawonn2024domain} shows a significant reduction in training time while also maintaining the classification accuracy of the non-parallelized approach. 
In this study, the practicability of the approach from~\cite{klawonn2024domain} will be further tested for a clinical, real-world dataset of human lung CT scans.
 Additionally, the present study comprises the following novelties.  Unlike~\cite{klawonn2024domain}, this study applies the model parallel training to a densely connected neural network (DenseNet)~\cite{huang2017densely} model for the first time while utilizing advanced data preprocessing and data augmentation techniques; cf. sec. \ref{sec:data_pre}, 
 with the aim to enhance the classification and generalization performance of the neural networks with respect to the considered multi-center multi-vendor dataset.
 The overall structure of the experimental design of this study is presented in Figure~\ref{fig:ex_design}.

\section{Materials and methods}
\label{sec:mat_meth}

In this study, we implemented and compared different 2D and 3D neural network models to address the binary classification of chest CT scans with and without signs of COVID-19-related pneumonia. 
We compare classification performance in terms of training, validation, and test accuracy, as well as training times for different CNN-based global neural network models with the classification performance and training time of a respective model parallel approach originally introduced in~\cite{klawonn2024domain}. 
Regarding the neural network architecture, we consider different CNN models that have been used successfully in medical image recognition, that is, a DenseNet121 model~\cite{huang2017densely}, as well as a 2D residual network (ResNet) model~\cite{he2016deep} and a 3D CNN model~\cite{tran2015learning}. 

\subsection{Dataset}
\label{sec:data}
 In this retrospective study, we collected a multi-center, multi-vendor dataset consisting of n=300 
chest CT scans of patients with COVID-19 (n=150) and non-COVID-19 pneumonia (nCP) (n=150) from three different university hospitals in Germany ~\cite{jaiswal2025, meng2023} Cologne (COVID-19: age 59.7±14.1, 52\% males; nCP: age 59.6±18.9, 56\% males), Frankfurt (COVID-19: age 58.5±13.9, 84\% males; nCP: age 59.9±13.3, 64\% males), Heidelberg (COVID-19: age 56.9±15.6, 68\% males; nCP: age 58.9±15.4, 68\% males). Informed consent was waived in this IRB-approved study (Cologne: 20-1676, Frankfurt: 20-719, and Heidelberg: S-293/2020). CT scans with pulmonary infiltration, and a positive RT-PCR test within 48 h before the CT examination are included in COVID-19 class. CT scans with inflammatory infiltrations, and an additional negative RT-PCR test after January 2020 are included in the nCP class. Cases with pneumonia caused by both viral and bacterial pathogens are included in nCP class.

\subsection{Data preprocessing}
\label{sec:data_pre}

\begin{figure}[ht]
\centering

\begin{subfigure}[b]{0.32\textwidth}
    \centering
    \includegraphics[height=3.35cm]{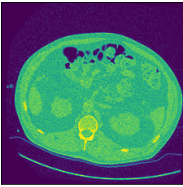}
    \caption{Original image.}
\end{subfigure}
\hfill
\begin{subfigure}[b]{0.32\textwidth}
    \centering
    \includegraphics[height=3cm]{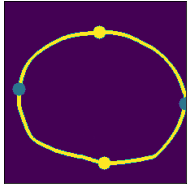}
    \caption{Biggest outer contour, extreme points.}
\end{subfigure}
\hfill
\begin{subfigure}[b]{0.32\textwidth}
    \centering
    \includegraphics[height=3.35cm]{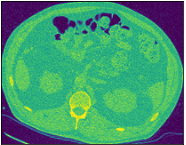}
    \caption{Cropped image.}
\end{subfigure}

\caption{Exemplary visualization of automatic body cropping applied to all CT scans.}
\label{fig:body_cropping}

\end{figure}

In order to enhance the generalization properties of all neural network models, different steps of data preprocessing and data augmentation were performed on both the training data and the validation data~\cite{jaiswal2025}. 
All data was transformed using the following steps: resampling to $224\times224$ pixels in the plane and $64$ pixels in the $z$-direction, normalization using min-max scaling, and intensity clipping to $(-1000,400)$ HU. This interval was selected to emphasize lung parenchymal attenuation while suppressing structures with substantially higher density.
Additionally, automatic body cropping was applied to all data by automatically finding the outer contours and extreme points of the human body; see Figure~\ref{fig:body_cropping} for an exemplary visualization. 
During training, the following data augmentation techniques were applied: Random rotation by angle randomly chosen from $[-20, -10, -5, 5, 10, 20]$ degree, cropping or padding by first enlarging every CT slice with $14\times14$ black pixels and subsequently randomly cropping it back to $224\times224$ pixels, and horizontal or vertical flipping.

\subsection{Neural network architectures}
\label{sec:nn}

 In this study, we considered the DenseNet121 model ~\cite{huang2017densely},  ResNet20~\cite{he2016deep}, and 3D CNN architectures~\cite{tran2015learning}. 
 DenseNet121 is a densely connected convolutional neural network model that consists of $121$ layers organized into four blocks. Each of these four blocks has separate pooling layers. It has been successfully used in different medical image classification tasks~\cite{singh2021densely,wang2019ensemble}. 
 ResNet20 is a 2D ResNet with $20$ layers. 3D CNN performs three-dimensional convolutions and $4$ blocks of convolutional layers with $32, 64, 128, 128$ filters. 
 
  The ResNet20 and the DenseNet121 receive each CT volume as a $224\times224\times64$ tensor, interpreting the $64$ axial slices as $64$ input channels, such that all residual blocks apply $3\times3$ in-plane convolutions that jointly integrate cross-slice information via channel-wise filtering.
The resulting feature maps are subsequently aggregated using global average pooling, yielding a compact representation of the entire volume that is passed to a fully connected softmax layer for final binary classification. 
  An overview of all three tested models is given in Table~\ref{tab:models}. 
 
 Given that the parallel model training approach from~\cite{klawonn2024domain} has, so far, exclusively been developed for convolutional neural network models, in this study, we also consider neural networks based on convolutional layers with and without skip connections.

\begin{table}[ht]
    \centering
   \scalebox{0.85}{
    \begin{tabular}{r|lll}
     & 2D ResNet & 2D DenseNet121 & 3D CNN \\\hline
    dimension & 2D & 2D & 3D \\
    no. layers/filters & 20 & 121 & [32,64,128,128] \\
    filter size & $3\times3$ & $3\times3$ & $3\times3\times3$ \\
    no. trainable params & 278\,674 & 7\,155\,266 & 755\,354\\
    activation fct. & ReLU & ReLU & ReLU \\
    input & $224\times224$ & $224\times224$ & $224\times224\times64$ \\
    initialization & ImageNet~\cite{he2016deep} &  ImageNet~\cite{huang2017densely} & He initialization~\cite{he2015delving} \\
    \end{tabular}
    }
    \caption{Overview of the three neural network models included in this study for the classification of COVID-19.}
    \label{tab:models}
\end{table}

\subsection{Model parallel training method}
\label{sec:model_parallel}

This section details the model parallel training approach~\cite{klawonn2024domain} for neural network-based image classification models. 
It decomposes a global image recognition model as a CNN into smaller models or subnetworks, respectively, which can be trained faster and in parallel. Subsequently, the approach reunites the resulting separate image classifications into a final decision by training and evaluating a small dense feedforward neural network (DNN) in a second phase. 

In more detail, the approach of~\cite{klawonn2024domain} assumes that we have two- or three-dimensional image data with $H\times W$ pixels or $H\times W\times D$ voxels, respectively.
With respect to the experiments presented in section~\ref{sec:results}, we will always present results for a \textit{global} CNN of a chosen type and compare the performance with the corresponding model parallel training approach, defined as CNN-DNN in~\cite{klawonn2024domain}. 

 For the definition of the CNN-DNN model of~\cite{klawonn2024domain}, we first decompose the input images into a finite number of $N \in \mathbb{N}$ smaller subimages.
 Let us note that, for the general case of three-dimensional voxel data with $H\times W\times D$ voxels, the input data can, in principle, be decomposed in all three dimensions, resulting in smaller image-based input data of size $H_i\times W_i\times D_i, \ i=1,\ldots, N$.
 For each of these subimages, we train proportionally smaller CNNs that operate exclusively on certain subimages of all input data. We refer to these smaller CNNs as \textit{local} CNNs for the remainder of this paper. These local CNNs can all be trained independently of each other and in parallel on different GPUs. 
 With the aim to provide a fair comparison between the global CNN and the CNN-DNN approach, the local CNNs are always of the same type as the global CNN (that is, residual networks, densely connected, etc.) but differ in the number of channels of the feature maps, the number of neurons within the fully connected layers, and the number of input nodes. All named layers are proportionally smaller than for the global CNN, corresponding to the reduced input dimension of the decomposed image input data. 
As output data of the local CNNs, we obtain $N$ different probability distributions with respect to the given image classification problem, each related to a local decision exclusively based on the information extracted from the local subimages. 

\begin{figure}[ht]
    \centering
    \includegraphics[width=0.9\textwidth]{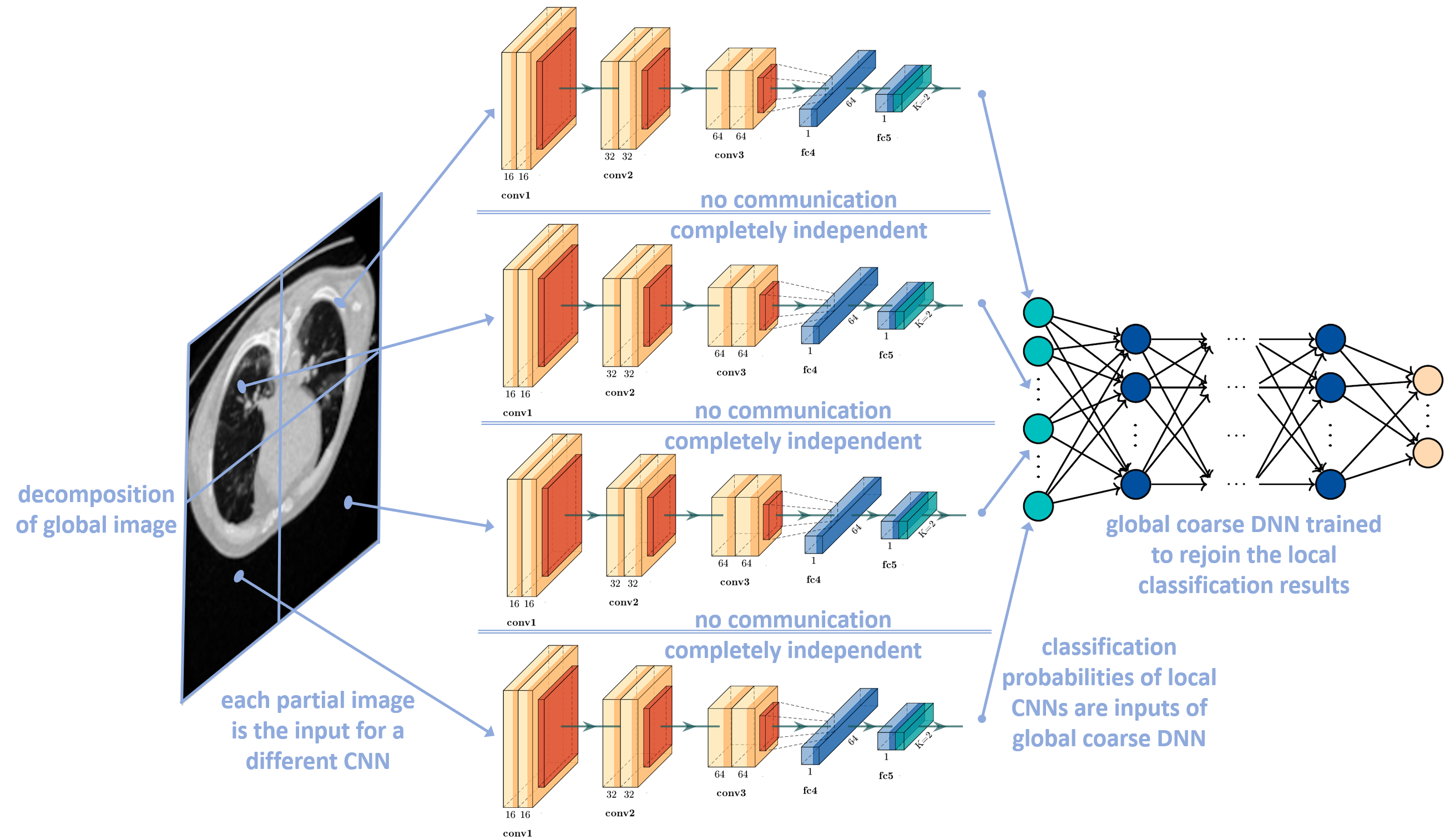}
    \caption{Visualization of the model parallel CNN-DNN approach as introduced in~\cite{klawonn2024domain} based on a spatial decomposition of the input image data into smaller subimages. 
    Figure adapted from~\cite[Fig. 4]{klawonn2024domain}.}
    \label{fig:CNN-DNN}
\end{figure}

In order to reunite these separate probability distributions into a final decision with respect to the classes of the considered image classification problem, in a second phase, a DNN is trained to compute an automatically optimized nonlinear combination; see~\cite{klawonn2024domain} for more details. 
An exemplary visualization of the implemented CNN-DNN model is shown in Figure~\ref{fig:CNN-DNN}.

\subsection{Implementation details and evaluation}
\label{sec:impl}

All networks have been implemented using Python 3.6 and TensorFlow-GPU 2.5~\cite{tensorflow2015-whitepaper}. For the body cropping in the data preprocessing, python's opencv package has been used. For the Floating Point Operation 
measurements of the ML models, we have used the tensorflow/keras model profiler. 
To train all the models, the Adam optimizer~\cite{kingma2014adam} and adaptive scaling of the learning rate on plateaus with a patience of $10$ epochs was used. Early stopping was implemented with patience of $15$ epochs~\cite{prechelt2002early}. 

For all experiments, we used a GPU cluster with 8 NVIDIA V100 GPUs. For the model parallel training method, we equally distributed the training of the local CNNs to the 8 GPUs. More precisely, the training of the $k$-th local CNN has been assigned to GPU with index=$k$ mod $8$. Subsequently, we have trained the DNN to automatically combine the predictions of the local networks on one GPU.

 Besides testing different neural network architectures, we compared the model parallel approach~\cite{klawonn2024domain} to the global network trained on the full CT scans without decomposition as a baseline.
 The dataset has been split into $80\%$ training, $10\%$ validation, and $10\%$ test data.
Model performance was evaluated on the independent test set using classification accuracy (ACC), receiver operating characteristic (ROC) curves, and the area under the ROC curve (AUC).
Time saving was computed from the GPU training times of the global and model-parallel approaches using three network architectures. For both approaches, total training time was measured until training termination by early stopping.

\section{Results}
\label{sec:results}

\begin{table}[ht]
    \centering
    \scalebox{0.86}{
    \begin{tabular}{l|rrr|rr}
    classification model & train acc & val acc & test acc  & train time & TtEval \\\hline
       global model & 0.77 & 0.71 & 0.71  & 02 h 05 min 57 s & 1.207 s \\
       CNN-DNN, $2\times2\times1$ & 0.88 & 0.75 & 0.74  & 35 min 13 s & 0.397 s  \\
       CNN-DNN, $4\times4\times1$ & 0.89 & 0.77 & 0.76  & 28 min 25 s & 0.231 s  \\
       CNN-DNN, $4\times4\times2$ & 0.87 & 0.75 & 0.75  & 19 min 12 s & 0.201 s
    \end{tabular}
    }
    \caption{Performance analysis for different classification models based on \textbf{DenseNet121} for medical image recognition of COVID-19 in chest CT scans (global model). As a benchmark, we always consider the global model, that is, one DenseNet121 model that operates on the entire chest CT scans. We show comparative results for the model parallel CNN-DNN approach with different numbers of subimages and local networks, respectively, cf. section~\ref{sec:model_parallel}. For all tested classification models, we report the classification accuracy on the training data (train acc), on the validation data (val acc), and on the independent test set (test acc). Additionally, we compare the training time (train time) and average time for evaluation (TtEval) of the different network models. 
    Times are reported in hours (h), minutes (min), and seconds (s).}
    \label{tab:Densenet121}
\end{table}

\begin{table}[ht]
    \centering
    \scalebox{0.87}{
    \begin{tabular}{l|rrr|rr}
    classification model & train acc & val acc & test acc  & train time & TtEval \\\hline
        global model   & 0.70 & 0.69 & 0.68  & 45 min 08 s & 0.804 s  \\
       CNN-DNN, $2\times2\times1$  & 0.68 & 0.66 & 0.63  & 10 min 08 s & 0.370 s \\
       CNN-DNN, $4\times4\times1$ & 0.71 & 0.66 & 0.66  & 04 min 37 s & 0.190 s \\
       CNN-DNN, $4\times4\times2$ & 0.72 & 0.66 & 0.65  & 02 min 23 s & 0.122 s 
    \end{tabular}
    }
    \caption{Performance analysis for different classification models based on a \textbf{2D ResNet20} for medical image recognition of COVID-19 in chest CT scans. See Table~\ref{tab:Densenet121} for the column labeling.}
    \label{tab:2d_ResNet}
\end{table}

\begin{table}[ht]
    \centering
    \scalebox{0.88}{
    \begin{tabular}{l|rrr|rr}
    classification model & train acc & val acc & test acc  & train time & TtEval \\\hline
       global model   & 0.72 & 0.70 & 0.69  & 15 h 13 min 19 s & 0.603 s \\
       CNN-DNN, $2\times2\times1$  & 0.87 & 0.75 & 0.74  & 51 min 43 s & 0.185 s  \\
       CNN-DNN, $4\times4\times1$ & 0.88 & 0.77 & 0.75  & 43 min 07 s & 0.097 s  \\
       CNN-DNN, $4\times4\times2$ & 0.84 & 0.74 & 0.74  &  28 min 50 s & 0.058 s
    \end{tabular}
    }
    \caption{Performance analysis for different classification models based on a \textbf{3D CNN} for medical image recognition of COVID-19 in chest CT scans. See Table~\ref{tab:Densenet121} for the column labeling.}
    \label{tab:3d_ResNet}
\end{table}

\subsection{Performance of the global models}
\label{sec:results_glob}

 The performance of the global models trained on the entire CT scans are presented in Tables ~\ref{tab:Densenet121},~\ref{tab:2d_ResNet}, and~\ref{tab:3d_ResNet}. In total, $3$ different global models have been trained. DenseNet121~\cite{huang2017densely} trained as one global model results in a validation and test accuracy of approximately $71\%$ which is only slightly lower than the observed training accuracy of $77\%$. 
The ResNet20 results in lower classification accuracies for both, the training data as well as the test and validation data than the DenseNet121 with approximately $70\%$ training accuracy and $68\%$ validation and test accuracy (see Table ~\ref{tab:2d_ResNet}). 
The ResNet20 is a smaller model than DenseNet121, which results also in faster training times. Given the lower classifications accuracy also with respect to the training data, the ResNet20 might not be complex enough for the given dataset which consists of a relatively small number of samples for both classes, that is, $n=150$ and $nCP=150$. 

For the 3D CNN model, the classification accuracy for the validation and test data is only slightly lower than for DenseNet121 (see Table~\ref{tab:3d_ResNet}). 
Hence, with respect to a possibly high classification accuracy, a three-dimensional CNN can be an effective alternative to the relatively large DenseNet model. However, it could be that a larger data set than available for this study is necessary to efficiently train a 3D CNN with a large number of trainable parameters. 

\begin{figure}[h!]
\centering
\begin{subfigure}[t]{0.47\textwidth}
\includegraphics[width=0.7\textwidth]{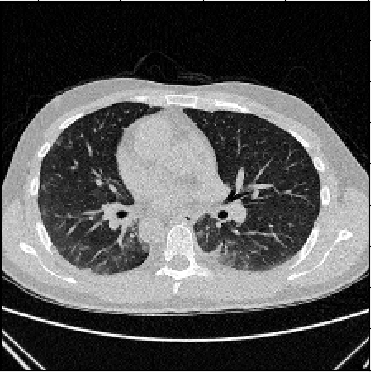}
\caption{COVID-19, correctly classified scan by DenseNet121.}
\end{subfigure}
\begin{subfigure}[t]{0.47\textwidth}
\includegraphics[width=0.7\textwidth]{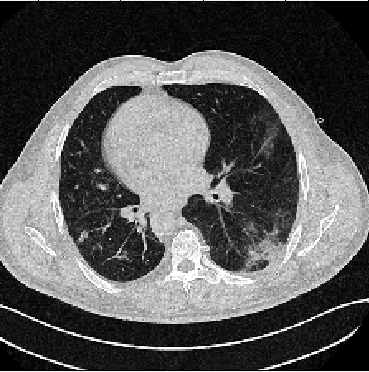}
\caption{nCP, incorrectly classified scan by DenseNet121.}
\end{subfigure}

\begin{subfigure}{0.47\textwidth}
\includegraphics[width=0.7\textwidth]{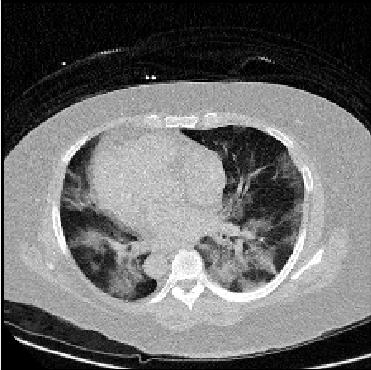}
\caption{COVID-19, correctly classified scan by 2D ResNet20.}
\end{subfigure}
\begin{subfigure}{0.47\textwidth}
\includegraphics[width=0.7\textwidth]{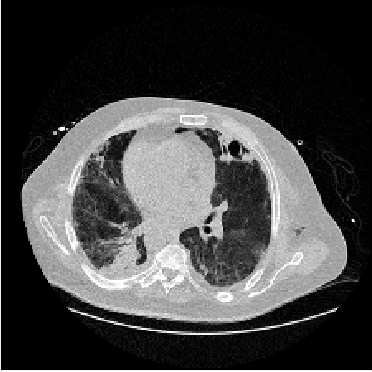}
\caption{nCP, incorrectly classified scan by 2D ResNet20.}
\end{subfigure}

\begin{subfigure}[t]{0.47\textwidth}
\includegraphics[width=0.7\textwidth]{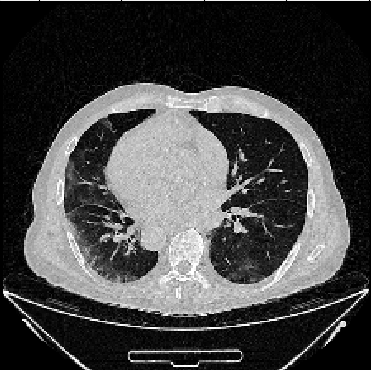}
\caption{COVID-19, correctly classified scan by 3D CNN.}
\end{subfigure}
\begin{subfigure}[t]{0.47\textwidth}
\includegraphics[width=0.7\textwidth]{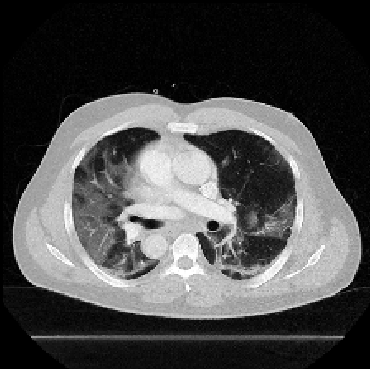}
\caption{nCP, incorrectly classified scan by 3D CNN.}
\end{subfigure}
\caption{Examples for correctly and incorrectly classified CT scans of all three tested global network models. (a) The CT scan is correctly classified as Covid by \textbf{DenseNet121}. (b) The CT scan is incorrectly classified as nCP by \textbf{DenseNet121}.
(c) The CT scan is correctly classified as Covid by \textbf{2D ResNet20}. (d) The CT scan is incorrectly classified as nCP by \textbf{2D ResNet20}.
(e) The CT scan is correctly classified as Covid by \textbf{3D CNN}. (f) The CT scan is incorrectly classified as nCP by \textbf{3D CNN}.
}
\label{fig:class_ex}
\end{figure}

\subsection{Performance of the model parallel approach}
\label{sec:results_parallel}

In this section, we investigate the performance of the model parallel training for all three network models as introduced above. In total, $9$ different CNN-DNN models using parallelization have been trained.
In Table~\ref{tab:Densenet121}, we observe that for the DenseNet121, the model parallel training, that is, the CNN-DNN approach, leads to enhanced classification accuracies for the training as well as the validation and test data, for all tested decompositions of the CT scans into smaller subimages. This indicates the decomposition of the input data seems to work quite well for the given dataset and the considered recognition of COVID-19 in chest CT scans and the DenseNet model. Additionally, the training time as well as the inference time or average time for evaluation is reduced drastically when comparing the CNN-DNN model to the respective global model. For the case of decomposing the CT scans into $2\times 2\times 1$ subimages, the training time is already reduced by a factor of approximately $3.6$, whereas it is reduced by a factor of approximately $6.6$ for $4\times 4\times 2$ subimages. 

For the ResNet20, the CNN-DNN model shows lower classification accuracies than the parallel trained DenseNet121 for all three split datasets; see Table~\ref{tab:2d_ResNet}. However, let us recall that also the global ResNet20 performs worse than the global DenseNet121 which could be due to the reduced complexity of the model architecture and the lower number of trainable parameters. 
Whereas for ResNet20, the classification accuracies of the CNN-DNN approach are $2$-$3\%$ lower than for the globally trained model, the training time could be reduced by a factor of approximately $4.5$ to $18.8$ for the different tested decompositions.

\begin{figure}[ht]
\centering
\begin{subfigure}[t]{0.32\textwidth}
\includegraphics[width=0.99\textwidth]{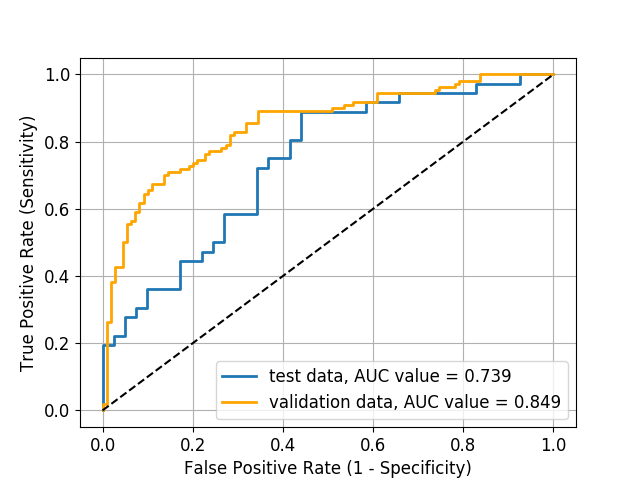}
\caption{DenseNet121, global.}
\end{subfigure}
\begin{subfigure}[t]{0.32\textwidth}
\includegraphics[width=0.99\textwidth]{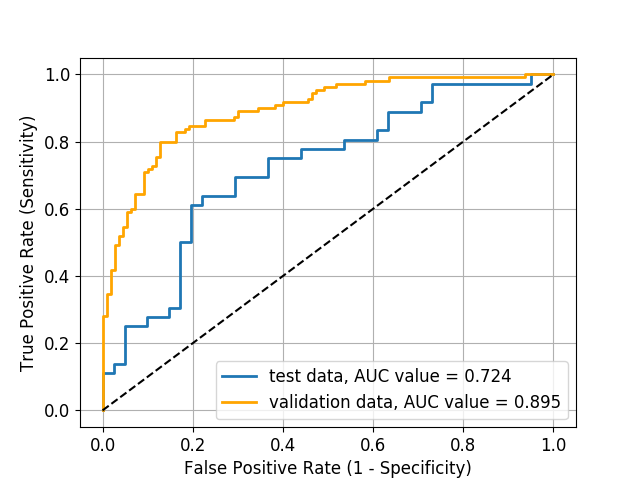}
\caption{2D ResNet20, global.}
\end{subfigure}
\begin{subfigure}[t]{0.32\textwidth}
\includegraphics[width=0.99\textwidth]{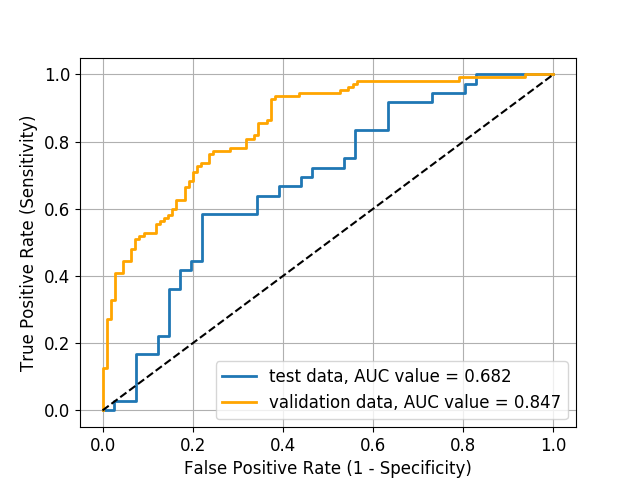}
\caption{3D CNN, global.}
\end{subfigure}

\begin{subfigure}[t]{0.32\textwidth}
\includegraphics[width=0.99\textwidth]{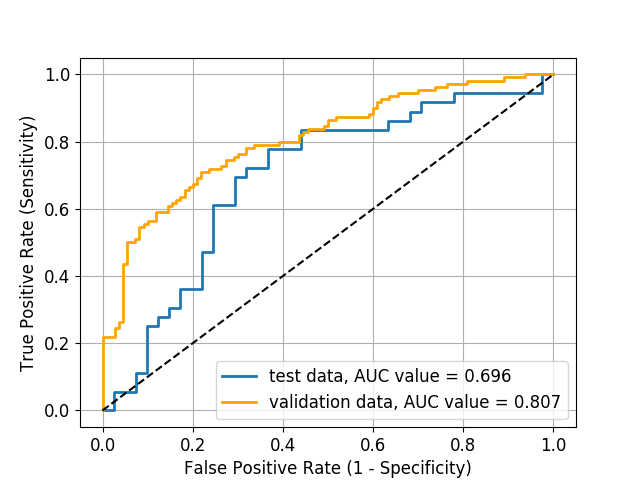}
\caption{DenseNet121, $2\times2\times1$.}
\end{subfigure}
\begin{subfigure}[t]{0.32\textwidth}
\includegraphics[width=0.99\textwidth]{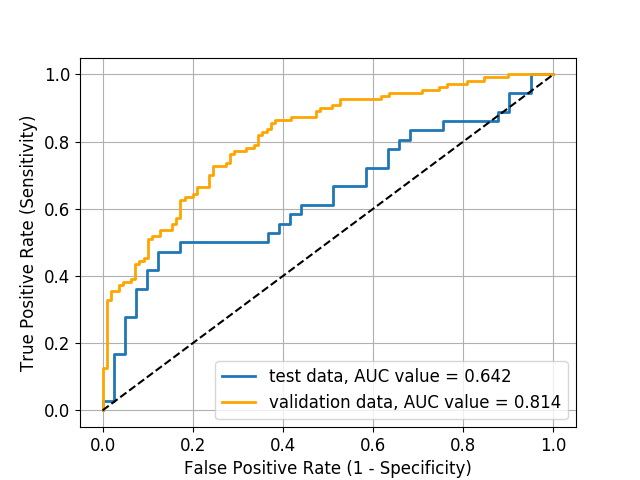}
\caption{2D ResNet20, $2\times2\times1$.}
\end{subfigure}
\begin{subfigure}[t]{0.32\textwidth}
\includegraphics[width=0.99\textwidth]{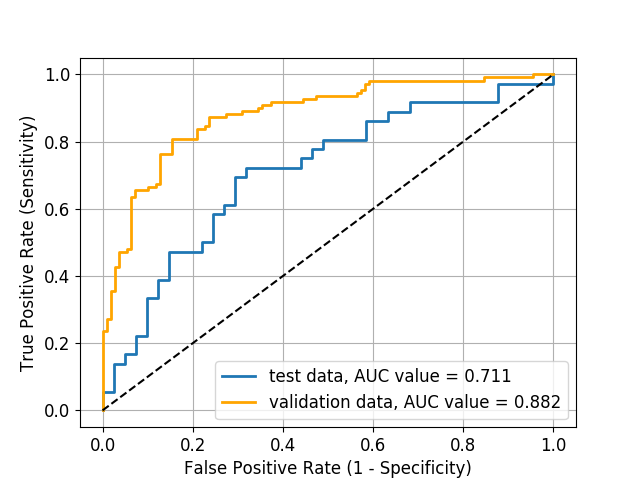}
\caption{3D CNN, $2\times2\times1$.}
\end{subfigure}

\begin{subfigure}[t]{0.32\textwidth}
\includegraphics[width=0.99\textwidth]{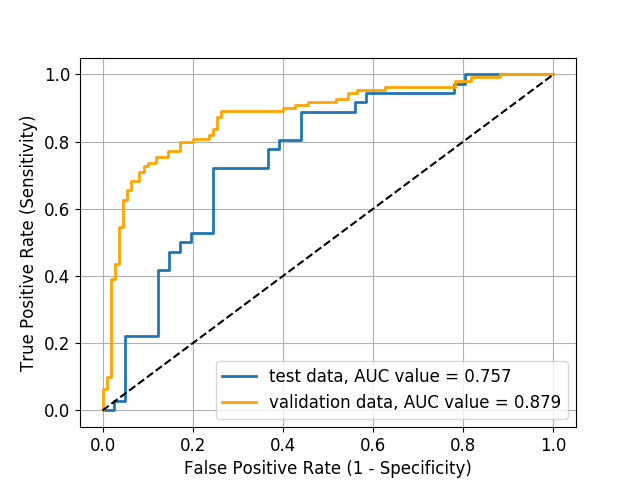}
\caption{DenseNet121, $4\times4\times1$.}
\end{subfigure}
\begin{subfigure}[t]{0.32\textwidth}
\includegraphics[width=0.99\textwidth]{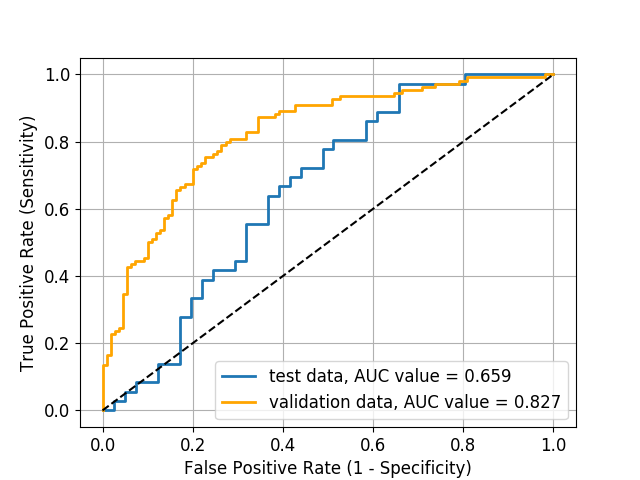}
\caption{2D ResNet20, $4\times4\times1$.}
\end{subfigure}
\begin{subfigure}[t]{0.32\textwidth}
\includegraphics[width=0.99\textwidth]{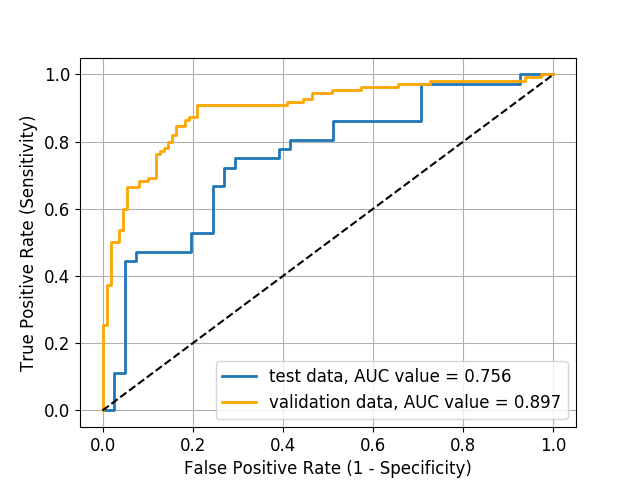}
\caption{3D CNN, $4\times4\times1$.}
\end{subfigure}

\begin{subfigure}[t]{0.32\textwidth}
\includegraphics[width=0.99\textwidth]{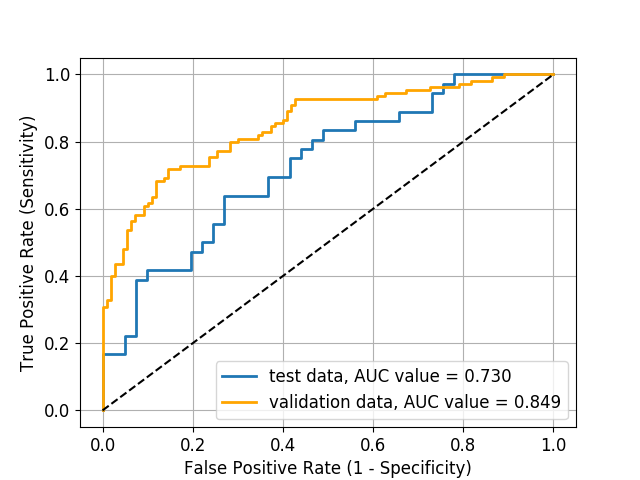}
\caption{DenseNet121, $4\times4\times2$.}
\end{subfigure}
\begin{subfigure}[t]{0.32\textwidth}
\includegraphics[width=0.99\textwidth]{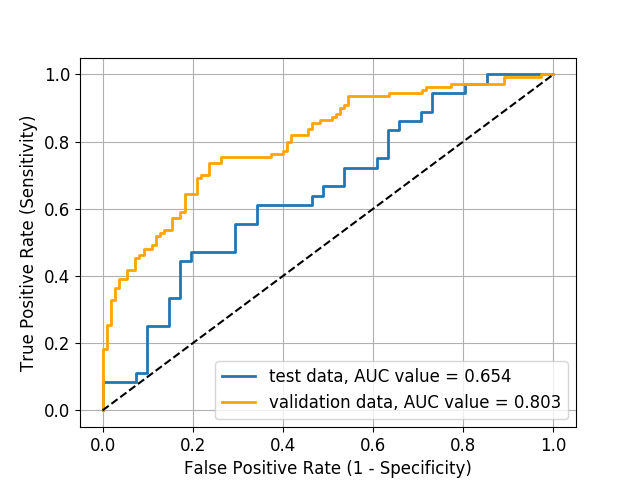}
\caption{2D ResNet20, $4\times4\times2$.}
\end{subfigure}
\begin{subfigure}[t]{0.32\textwidth}
\includegraphics[width=0.99\textwidth]{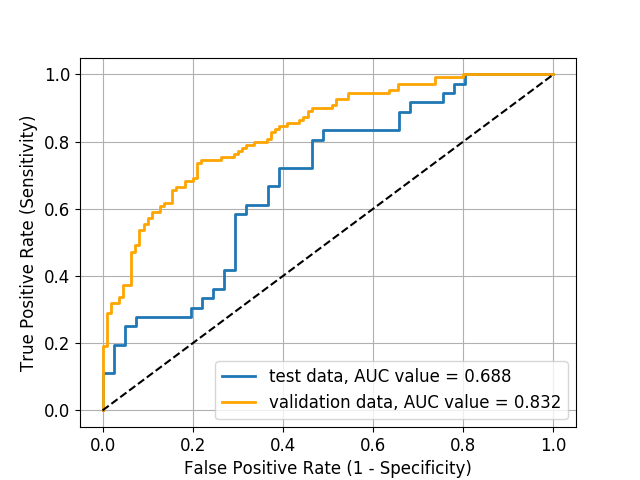}
\caption{3D CNN, $4\times4\times2$.}
\end{subfigure}
\caption{ROC curves for DenseNet121 ((a), (d), (g), (j)), 2D ResNet20 ((b), (e), (h), (k)), and 3D CNN ((c), (f), (i), (l)) for the validation and test data for global models and $2\times2\times1$, $4\times4\times1$, and $4\times4\times2$ subdomains.}
\label{fig:roc_curves}
\end{figure}

Finally, from Table~\ref{tab:3d_ResNet}, we observe that for the three-dimensional CNN the parallelization approach works quite well again. Here, similarly as for the DenseNet model, the parallelization in form of a localization of the input data leads to enhanced classification accuracy values, in particular for the validation and test data. 
Additionally, for the 3D CNN model, the training time can be reduced by a factor of $17$ to $31$ using the tested model parallel training approach. 
Examples for correctly and incorrectly classified CT scans for all three tested network architectures are summarized in Figure~\ref{fig:class_ex}. 
Additionally, we provide the ROC (receiver operating characteristics) curves for the tested network models in Figure~\ref{fig:roc_curves}.

\section{Discussion and conclusion}
\label{sec:disc}

In recent years, the potential of AI methods to assist radiologists in detection, classification, and, more general, disease diagnosis tasks, has strongly grown and emerged to a wide research field and is now rapidly making its way into clinical practice. This evolution stresses the importance of the development of powerful AI and ML methods for this purposes as well as the necessity for efficient implementation pipelines to successfully train and evaluate the underlying ML models. 
While many approaches exist and have been tested for a data parallel training of different classification models (see also~\cite{ben2019demystifying} for a methodological overview), in the current study, a model parallel training strategy for the classification of two- or three-dimensional image data is tested for the exemplary application of COVID-19 diagnosis based on chest CT scans. 
 
For all three tested models, the training time of the respective model as well as the average inference time was substantially reduced by the parallel approach. We have observed the highest potential to reduce the training time for the 3D CNN model. This is in compliance with the results from~\cite{klawonn2024domain} where also the highest reduction in training time was observed for a CNN with three-dimensional filters. 

Furthermore, the model parallel CNN-DNN model showed improved classification accuracies on the validation and test data set for the DenseNet121 and the 3D CNN model. Hence, the decomposition of the input data into smaller subimages and the separate training of local smaller networks, that are automatically recombined with a small DNN, tends to also help the accurate identification of signs of COVID-related pneumonia in the considered CT scans. Similar observations have also been made in~\cite{gu2022decomposition} and~\cite{klawonn2024model}, where also the decomposition and composition of deep convolutional neural networks in combination with subnetwork transfer learning helps to increase the accuracy of the considered image classification. Additionally, in~\cite{klawonn2024model}, it was also shown that the training of a small DNN to combine the local classification of smaller networks trained in parallel outperforms a simple majority voting or an averaging of the local probability values. 

The best overall validation and test accuracy were obtained for DenseNet121 and 3D CNN. The ResNet20 model performed worst in both global and parallel settings. Hence, a well-performing global neural network model must be chosen as a basis for a successful parallel model. Overall moderate accuracy obtained is attributed to the classification design (COVID-19 vs. nCP) and is in accordance with the previous studies~\cite{jaiswal2025, meng2023}. Distinguishing between COVID-19 and nCP is inherently a challenging task due to similar patterns being exhibited. Nevertheless, the diagnostic approach was found to be helpful as a support tool~\cite{meng2023}. COVID-19 pandemic has motivated development of networks, infrastructure, and processes to streamline pandemic data collection, availability, and timely analysis to prepare for a proactive response all around the world~\cite{babady2022building, martinez2025cloud, xia2024canada, heyder2023german}. Our approach complements the pandemic preparedness effort.

This study considers the COVID-19 pandemic as an example use case. During the early COVID-19 pandemic, CT was the recommended way to diagnose COVID pneumonia ~\cite{KWEE2020} before the virus was even identified and sequenced, and mass production of SARS CoV swab tests was started. The first antigen test was approved by the FDA about half a year after the pandemic started (https://www.fda.gov/news-events/press-announcements/coronavirus-covid-19-update-fda-authorizes-first-antigen-test-help-rapid-detection-virus-causes). Before that, RT-PCR testing (time consuming and expensive, limited capacities) and chest CT were used. Radiologists helped bridge the time until swab tests were available. However, the stress on radiology CT departments increases substantially. Therefore, in future pandemics, AI must be the tool to enable for early mass testing with CT and fast retraining caused by changing imaging pattern due to changing virus subclones during the pandemia.

 This study has certain limitations. First, the parallel CNN-DNN training approach has so far been applied exclusively to CNN-based architectures.  Its transfer to other architectures, for example, using transformer layers, will be a topic of future research. Second, the application was limited to binary classification. Validation on multi-class classification formulations, more centers, and other diseases is warranted before clinical implementation.

 In conclusion, the parallel CNN-DNN training approach has achieved consistent saving of computing time (training and inference) across different architectures with an often improved diagnostic accuracy. It efficiently utilizes the data as well as infrastructure. The CNN-DNN architecture is transferable to multiple classification networks. Further validation at different centers is needed before widespread clinical implementation.
 
\clearpage 

\section*{Conflict of interest} The authors declare no conflict of interest.

\section*{Acknowledgements} We gratefully acknowledge the use of the computational
facilities of the Center for Data and Simulation Science (CDS) at the University of
Cologne. This work has been supported by RACOON NUM 2.0 ``(FKZ: 01KX2121) and NUM 3.0'' (FKZ: 01KX2524).

\section*{Authors' contributions} JWH: Methodology, Implementation, Visualization, Formal Analysis, Writing – original draft, Writing – review \& editing. AJ: Data curation, Writing – original draft, Writing – review \& editing. PF: Data curation, Writing – review \& editing. DM:  Data curation. AG:  Data curation. RF: Data curation. AMB: Data curation. RS:  Data curation, Writing – review \& editing. JK:  Data curation, Writing – review \& editing. TP: Conceptualization, Writing – review \& editing. AK: Conceptualization, Methodology, Supervision, Writing – review \& editing.






\bibliographystyle{unsrt}
\bibliography{im_seg.bib}




\newpage 
\appendix

\section{Supplementary material}

\begin{figure}[h!]
    \centering
    \includegraphics[width=0.9\textwidth]{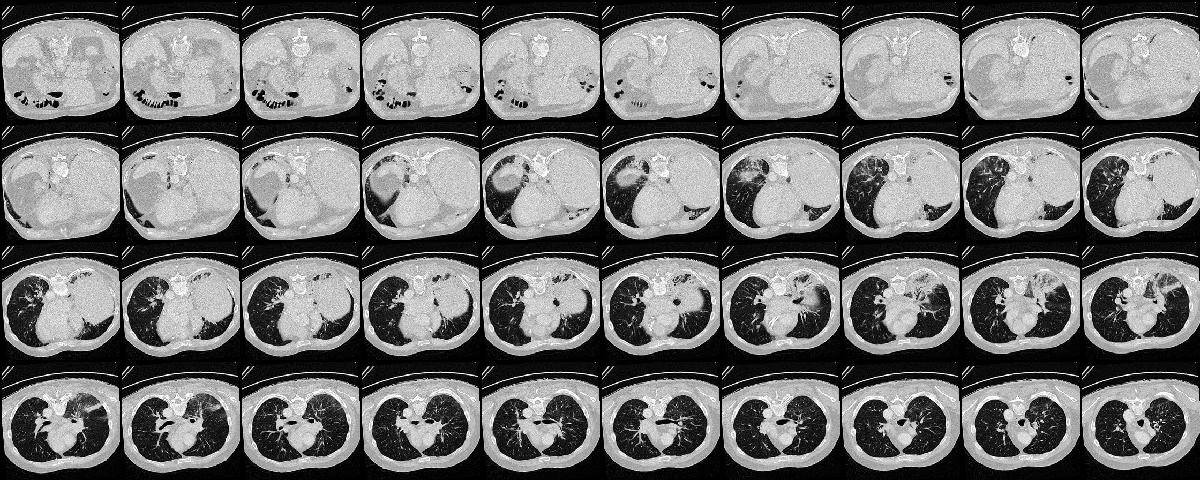}
    \caption{Exemplary slices from a chest CT scan from our dataset. All slices are scaled to $224\times 224\times 64$ voxels. }
    \label{fig:ct_examples}
\end{figure}


\end{document}